# Statistical Comparative Analysis of Semantic Similarities and Model Transferability Across Datasets for Short Answer Grading

Sridevi Bonthu*[1,4], S. Rama Sree[2], M. H. M. Krishna Prasad[3]

**Abstract:** Developing dataset-specific models involves iterative fine-tuning and optimization, incurring significant costs over time. This study investigates the transferability of state-of-the-art (SOTA) models trained on established datasets to an unexplored text dataset. The key question is whether the knowledge embedded within SOTA models from existing datasets can be harnessed to achieve high-performance results on a new domain. In pursuit of this inquiry, two well-established benchmarks, the STSB and Mohler datasets, are selected, while the recently introduced SPRAG dataset serves as the unexplored domain. By employing robust similarity metrics and statistical techniques, a meticulous comparative analysis of these datasets is conducted. The primary goal of this work is to yield comprehensive insights into the potential applicability and adaptability of SOTA models. The outcomes of this research have the potential to reshape the landscape of natural language processing (NLP) by unlocking the ability to leverage existing models for diverse datasets. This may lead to a reduction in the demand for resource-intensive, dataset-specific training, thereby accelerating advancements in NLP and paving the way for more efficient model deployment.

## 1. Introduction

Natural Language Processing (NLP) stands as a pivotal branch of artificial intelligence that focuses on the interaction between human language and computational systems [1]. At its core, NLP strives to bridge the gap between the complexities of human communication and the computational capabilities of machines [2]. By enabling machines to understand, interpret, and generate human language, NLP has found widespread applications across various domains, shaping the way we interact with technology and transforming industries ranging from communication and commerce to healthcare and entertainment [3]. Dataset-specific model training in NLP holds paramount importance in tailoring machine learning algorithms to the intricacies of particular domains, thereby enabling accurate predictions and insights. It acknowledges that data characteristics can vary substantially across different applications, making a customized approach essential [4]. This practice enhances model performance, yielding results that align more closely with real-world scenarios. Moreover, domain-specific models often outperform generic models, showcasing the value of focused training. However, dataset-specific training is not without challenges. Acquiring, curating, and annotating domain-specific data demands substantial resources and domain expertise. Overfitting can result from excessive model customization, impacting the model's generalization to new data [5]. The need for continuous adaptation and retraining as the domain evolves adds another layer of complexity.

Automated Short Answer Grading (ASAG) involves the automated assessment of student-authored responses through a comparison with reference answers [6]. State-of-the-art (SOTA) models in the ASAG field employ similarity metrics to make score predictions [7]. However, the availability of datasets for this specific task remains limited, despite a growing number of researchers introducing new datasets [8]. The process of starting from scratch and constructing models anew proves both resource-intensive and time-consuming. In response, a crucial need arises for investigating model transferability within this domain to assist researchers. In light of this need, our study focuses on addressing this challenge. We take a practical approach by selecting two established datasets, namely Mohler [9] and STSB [10], and subject them to a comparative analysis alongside a novel dataset called SPRAG [8]. This comparative examination aims to shed light on the extent to which models developed on established datasets can be effectively applied to a new and unexplored dataset. The primary aims of this study encompass:

- Conducting a comparative analysis between well-established and newly introduced datasets utilizing both contextual and non-contextual similarity metrics.

- Employing statistical tools such as the t-test and

[1] *Jawaharlal Nehru Technological University, Kakinada, India*
*ORCID ID : 0000-0002-1971-4965*
[2] *Aditya Engineering College, Surampalem, India*
*ORCID ID : 0000-0002-8771-6006*
[3] *Jawaharlal Nehru Technological University, Kakinada, India*
*ORCID ID : 0000-0003-2150-7807*
[4] *Vishnu Institute of Technology, Bhimavaram, India*
*ORCID ID : 0000-0002-1971-4965*
* *Corresponding Author Email: sridevi.db@email.com*

- Cohen's d to rigorously analyze the observed similarities.
- Offering valuable insights into the potential transferability of models across different datasets.

The subsequent sections of the paper are structured as follows: In Section 2, various techniques for dataset comparison are explored. Section 3 outlines the approach employed to conduct the study. The experimental outcomes are presented in Section 4, followed by their interpretation in Section 5. Lastly, Section 6 encapsulates the conclusions drawn from the study and outlines potential for future research.

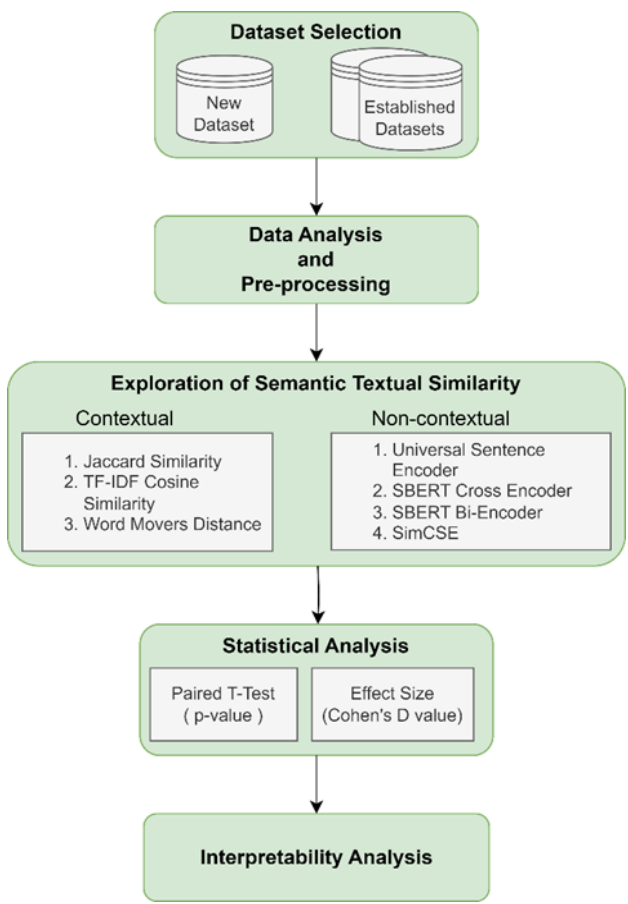

**Fig. 1.** Proposed framework for comparing the datasets.

## 2. Related Work

In this section, we present the diverse methodologies employed for dataset comparison. Araque et al. [11] conducted experiments utilizing semantic similarity metrics to assess the comparability of text columns within a dataset. Azarpanah et al. [12] elucidated the biases inherent in measuring similarity metrics. Bag S et al. [13] undertook experiments employing traditional similarity metrics, yielding favorable outcomes. Chiny et al. [14] applied TF-IDF based cosine similarity on sentence vectors to enhance recommendations for Netflix movie systems. Kusner et al. [15] demonstrated the superiority of distance metrics, such as Word Mover Distance, in comparison to word embeddings for finding similarities. Cer D et al. [16] introduced a Universal Sentence Encoder architecture that effectively produces word embeddings, which are instrumental in gauging similarity between textual inputs. Reimers et al. [17] introduced a twin Siamese architecture based on BERT for generating sentence embeddings. Thakur et al. [18] investigated SBERT under two settings: bi-encoder and cross-encoder. Both architectures exhibited robust performance in evaluating text similarities. Gao T et al. [19] proposed a straightforward contrastive learning approach for sentence embeddings. Collectively, the array of contextual and non-contextual similarity metrics discussed above can be harnessed to compare datasets effectively.

Statistical tests serve as powerful tools for quantitatively comparing datasets, aiding in uncovering meaningful insights and drawing reliable conclusions. These tests provide a systematic framework to assess whether observed differences between datasets are statistically significant or merely due to random chance. Hypothesis testing involves formulating a null hypothesis (H0) and an alternative hypothesis (H1). The null hypothesis assumes no significant difference between groups, while the alternative hypothesis posits a difference exists. By conducting statistical tests, we determine whether there's enough evidence to reject the null hypothesis in favor of the alternative [20]. The t-test is a widely used parametric test to compare means between two groups. It assesses whether the observed difference between sample means is statistically significant or likely due to random variability [21]. Effect size quantifies the magnitude of the difference between groups or conditions. It complements significance testing by providing a measure of practical or clinical significance [22]. Building upon the aforementioned research, we opt to conduct a comprehensive dataset comparison employing both contextual and non-contextual similarity metrics, as well as employing paired t-test and effect size calculations.

## 3. Methodology

This section outlines the approach used to compare the datasets, as depicted in the Fig. 1. The process involves a series of steps. It begins with the analysis and pre-processing of the selected datasets. Subsequently, semantic textual similarity is computed using both contextual and non-contextual similarity metrics. This is followed by conducting statistical analysis using a paired t-test along with the assessment of effect size. Lastly, the outcomes are examined and interpreted.

### 3.1. Datasets

This study employed three datasets for short answer grading: two well-established datasets (STSB [10], Mohler [9]) and a novel dataset (SPRAG [8]). Each dataset comprises pairs of sentences along with a numerical label indicating their similarity, ranging from 0 (least similar) to 5 (most similar). The STSB (Semantic Textual Similarity Benchmark) dataset is a widely acknowledged benchmark within the realm of NLP. It is commonly employed to assess the algorithms and the models designed to evaluate the semantic likeness. The Mohler dataset has gained popularity among the ASAG research community. On the other hand, SPRAG is a recently developed dataset centered around the domain of Python programming. While the STSB and Mohler datasets feature sentences written in natural English language, SPRAG includes certain keywords and symbols like def, del, elif, *, & # etc., integrated into its sentences.

**Table 1.** Top 20 common words in the three datasets considered

| Mohler | STSB | SPRAG |
|---|---|---|
| **function** 1490 | man 1540 | value 2245 |
| array 1137 | woman 993 | **function** 2036 |
| element 1007 | dog 739 | **datum** 1834 |
| **list** 811 | say 686 | python 1451 |
| node 628 | play 645 | **list** 1314 |
| pointer 486 | kill 592 | return 1124 |
| **datum** 465 | white 435 | code 1027 |
| **variable** 429 | black 391 | give 990 |
| member 368 | run 339 | argument 940 |
| type 357 | people 318 | method 817 |
| link 353 | percent 318 | operator 810 |
| **call** 336 | stand 284 | **variable** 764 |
| memory 323 | new 282 | program 754 |
| tree 322 | girl 275 | object 744 |
| right 311 | sit 273 | line 739 |
| store 303 | ride 254 | number 715 |
| class 299 | boy 251 | **call** 615 |
| operation 294 | year 249 | string 546 |
| point 266 | state 243 | keyword 546 |
| declare 266 | water 226 | statement 540 |

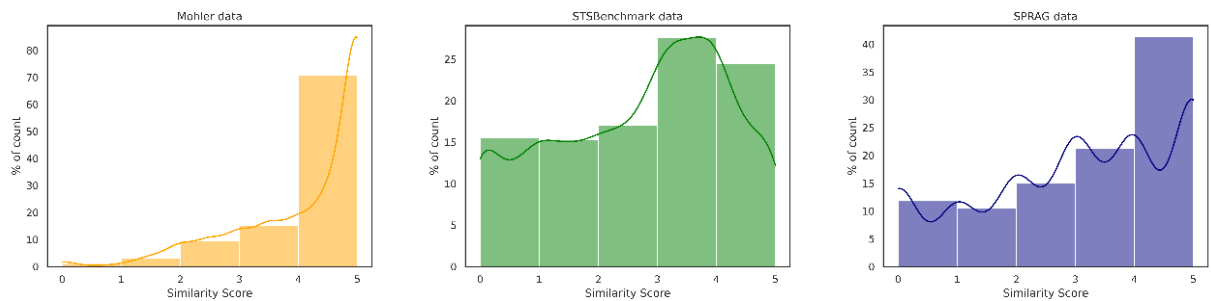

**Fig. 2.** Distribution of the similarity scores between the two sentences in the datasets.

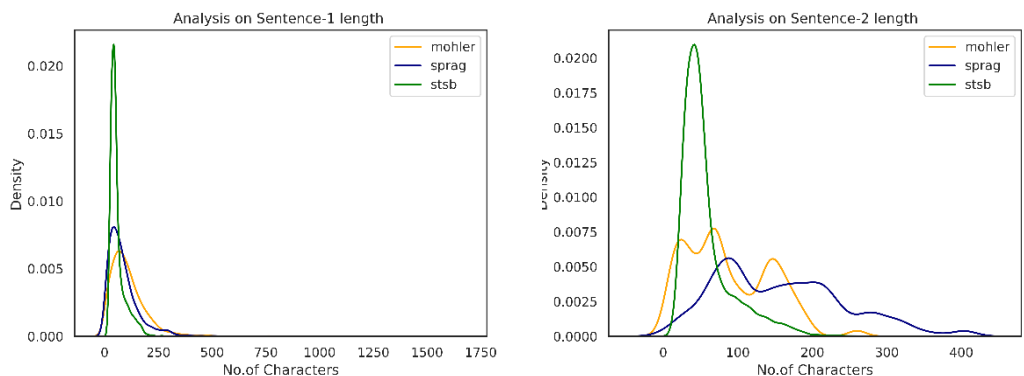

**Fig. 3.** Density plots of sentence lengths of Mohler, SPRAG and STSB datasets

The distribution of similarity scores for all datasets is visualized in the Fig. 2. Notably, the Mohler dataset exhibits a significant imbalance, with the majority of examples falling into label 5. In contrast, the STSB dataset showcases a well-balanced distribution compared to the other two datasets. Mohler's distribution suggests a higher likelihood of models being overfitted, whereas the STSB dataset shows a greater potential for generalizability. In the case of the SPRAG dataset, label 5 encompasses a substantial portion of the records. Simultaneously, the class distribution for the remaining five classes demonstrates a favourable balance. The density plots of lengths of sentence-1 and sentence-2 are shown in Fig. 3. The sentence lengths of STSB datasets are smaller when compared with SPRAG and Mohler. The top 20 common words from the datasets are tabulated in Table 1. The table also presents the frequency of occurrence every word next to it. In the SPRAG dataset, the top 10 words are noticeably more frequent and exclusively pertain to the programming domain. In contrast, this pattern does not hold true for the other two datasets, namely STSB and Mohler. There is a 20% match in Mohler and SPRAG whereas there is no intersection of the words in the STSB dataset with the other datasets.

### 3.2. Semantic Similarity metrics

Similarity algorithms are used to measure the similarity between words, phrases and documents [23]. These techniques play a crucial role in the applications like text summarization, document clustering, search engine ranking, and many more. Both contextual and non-contextual algorithms are employed in this work to measure the similarity [12]. Non-contextual similarity algorithms do not consider the surrounding context and rely solely on predefined lexical or linguistic properties to measure similarity. These algorithms often employ predefined features or metrics that capture specific linguistic characteristics of words or phrases. Contextual similarity algorithms take into account the surrounding context of words or phrases to determine their similarity. These algorithms consider the words that appear nearby and the relationships between them. Contextual similarity aims to capture the meaning and semantics of words based on their usage in a specific context. Contextual models, like word embeddings derived from neural networks, often excel at capturing these nuances.

### 3.2.1. Non-contextual Similarity algorithms

**Jaccard Similarity.** It is a simple and intuitive measure used to assess the similarity between two sets. It's often applied in text analysis for comparing the overlap between sets of words [24]. The Jaccard similarity measures the ratio of the size of the intersection of two sets to the size of their union. In the context of text analysis, it quantifies the overlap of terms between two documents, making it useful for tasks such as document clustering and content recommendation. Eq. (1) shows the measurement of Jaccard Similarity between two sets $S_1$ and $S_2$.

$$J(S_1, S_2) = \frac{|S_1 \cap S_2|}{|S_1 \cup S_2|} \quad (1)$$

**TF-IDF Cosine Similarity.** TF-IDF (Term Frequency-Inverse Document Frequency) cosine similarity measures the cosine of the angle between two vectors representing the term frequencies of words in a document as shown in equation [25]. The Eq. (2) gives the value of every term $i$ in document $j$, where $tf_{ij}$ is the number of occurrences of $i$ in $j$, $df_i$ is the number of documents containing $i$ and $N$ is the total number of documents. It's a widely used method for comparing the similarity between documents based on their term distributions. A cosine similarity between the obtained vectors $S_1$ and $S_2$ is calculated by using the formula shown in Eq. (3).

$$w_{ij} = tf_{ij} \times \log\left(\frac{N}{df_i}\right) \quad (2)$$

$$Cosine\_similarity(S_1, S_2) = \frac{S_1 . S_2}{\|S_1\|\|S_2\|}$$
$$= \frac{\sum_{i=1}^{n} S_{1_i} S_{2_i}}{\sqrt{\sum_{i=1}^{n} S_{1_i}^2} \sqrt{\sum_{i=1}^{n} S_{2_i}^2}} \quad (3)$$

TF-IDF assigns weights to terms based on their frequency in a document relative to their frequency across a corpus. Cosine similarity between TF-IDF vectors quantifies the similarity of two documents in terms of their content, often used in search engines and text classification.

**Word Mover Distance (WMD).** It is a distance metric that measures the "distance" between two documents in a word embedding space [26]. It considers the cost of transforming words from one document to another, capturing semantic similarities. The Euclidean distance $c(i,j)$ in embedding space of two words $x_i$ and $x_j$ is given by $\|x_i - x_j\|^2$. In WMD, $x_i$ and $x_j$ are from different documents and $c(i,j)$ is the travel cost from word $x_i$ to $x_j$. WMD similarity measure accounts for semantic relationships between words by modeling how words can be "moved" from one document to another. This makes it particularly useful for tasks that require understanding the semantic similarity of text, such as document summarization and machine translation.

### 3.2.2. Contextual Similarity algorithms

**Universal Sentence Encoder (USE).** It is a pre-trained deep learning model that encodes sentences into high-dimensional vectors capturing semantic meaning [16]. This encoder leverages a transformer architecture to generate fixed-length representations for sentences. By utilizing this encoder, we aim to measure the contextual similarity of sentences by computing the cosine similarity between their encoded vectors.

**SBERT Cross Encoder (SBERT CE).** It is based on the Sentence-BERT framework, extends the capability of encoding pairs of sentences [17]. It captures the relationship between two sentences and produces embeddings suitable for similarity comparison. Incorporating the SBERT Cross Encoder, we intend to analyze the contextual similarity between sentence pairs with a focus on cross-sentence relationships.

**SBERT Bi-Encoder (SBERT BiE).** It is another variation of Sentence-BERT that independently encodes each sentence in a pair. This approach aims to represent each sentence as a standalone embedding, facilitating a comparison that is specific to each individual sentence [17]. We utilize the SBERT Bi-Encoder to investigate the contextual similarity of sentences without explicitly considering their relationship to each other.

**SimCSE Supervised.** SimCSE (Simple Contrastive Learning of Sentence Embeddings) supervised is a technique that leverages contrastive learning to enhance sentence embeddings' discriminative power [19]. By incorporating this method, we aim to extract sentence embeddings that are not only contextually similar but also optimized for specific classification tasks. This approach aligns with our goal of evaluating contextual similarity in the context of supervised learning scenarios.

**Unsupervised SimCSE.** It is a variant of the SimCSE framework designed for unsupervised tasks. It focuses on learning sentence embeddings without explicit label information, which allows for a broader exploration of contextual similarity in scenarios where labeled data may be scarce [19]. Integrating Unsupervised SimCSE, we explore the contextual similarity of sentences in an unsupervised context.

### 3.3. Statistical Analysis

Statistical analyses serve as a mathematical approach to determining the extent of significance in the dissimilarity between two datasets. These datasets are subject to comparison through conventional descriptive statistical methods. This undertaking also involves the execution of the t-test, enabling the quantification of the effect size to provide a more concise representation of the variance between the datasets.

**The Paired T-test**, a statistical procedure, is specifically designed for evaluating the means of identical groups or items within distinct scenarios. In this instance, the null hypothesis is set as "No substantial disparity exists in the mean cosine similarity scores among datasets A, B, and C," while the alternative hypothesis states that "A significant difference exists in the mean cosine similarity scores among datasets." Given the three datasets—namely Mohler, STSB, and SPRAG—the t-test is carried out on dataset pairs such as (Mohler, STSB), (Mohler, SPRAG), and (SPRAG, STSB). The paired t-test is performed by using the Eq. (4).

$$t = \frac{\sum d}{\sqrt{\frac{n(\sum d^2) - (\sum d)^2}{n-1}}}, \quad (4)$$

where $\sum d$ is the sum of the differences.

**Effect size,** specifically Cohen's d, is a statistical measure that quantifies the magnitude of the difference between two groups or conditions in a study. It's used to express the practical significance or real-world impact of an observed effect, in contrast to just measuring statistical significance. Cohen's d is calculated by taking the difference between the means of the two groups and dividing it by the pooled standard deviation. The formula to find Cohen's d is shown in the Eq. (5), where $Mean_1$ and $Mean_2$ are the means of the two groups being compared, the pooled standard deviation is a weighted average of the standard deviations of the two groups.

$$cohen's\ d = \frac{Mean_1 - Mean_2}{Pooled\ Standard\ Deviation} \quad (5)$$

The resulting value of Cohen's d indicates the standardized effect size. A larger Cohen's d suggests a greater difference between the groups, while a smaller value indicates a smaller difference. Generally, Cohen's guidelines for interpreting the effect size are as follows:

Small effect size : $d \approx 0.2$

Medium effect size : $d \approx 0.5$

Large effect size : $d \approx 0.8$

Cohen's d helps researchers and analysts better understand the practical significance of the findings and provides additional insight beyond just determining whether the results are statistically significant.

## 4. Experiment Results

**Table 2.** Similarity scores obtained by the metrics on the pairs of datasets.

| Similarity Measure | (STSB, Mohler) | (STSB, SPRAG) | (Mohler, SPRAG) |
|---|---|---|---|
| **Jaccard** | $2.64 \times 10^{-19}$ | $5.55 \times 10^{-22}$ | $4.41 \times 10^{-01}$ |
| **TFIDF** | $1.31 \times 10^{-04}$ | $2.75 \times 10^{-15}$ | $1.09 \times 10^{-05}$ |
| **NegWMD** | $8.78 \times 10^{-21}$ | $1.49 \times 10^{-22}$ | $6.89 \times 10^{-01}$ |
| **USE** | $4.11 \times 10^{-09}$ | $1.42 \times 10^{-10}$ | $8.14 \times 10^{-01}$ |
| **SBERT CE** | $9.63 \times 10^{-13}$ | $2.88 \times 10^{-03}$ | $5.34 \times 10^{-08}$ |
| **SBERT BiE** | $1.72 \times 10^{-16}$ | $1.29 \times 10^{-03}$ | $4.30 \times 10^{-11}$ |

| SimCSE Supervised | $1.25 \times 10^{-14}$ | $8.69 \times 10^{-04}$ | $1.40 \times 10^{-08}$ |
| SimCSE Unsupervised | $3.20 \times 10^{-01}$ | $9.55 \times 10^{-05}$ | $5.22 \times 10^{-08}$ |

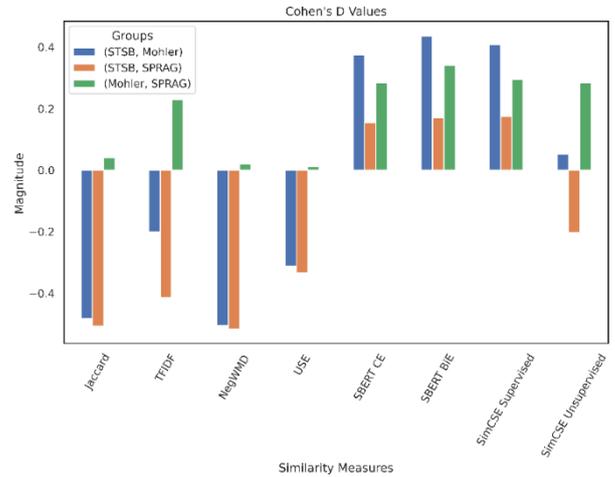

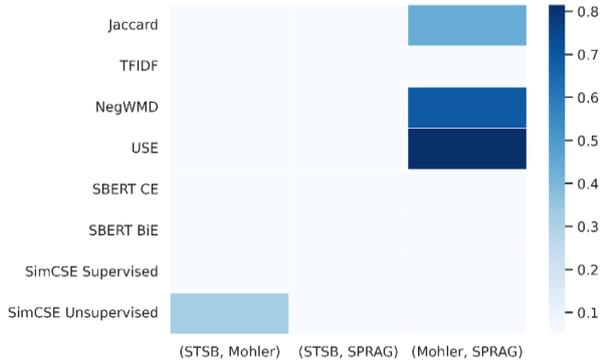

**Fig. 4.** heatmap of the similarity scores on pairs of datasets.

The three datasets are organized into distinct pairs, leading to the formation of three groups: (STSB, Mohler), (STSB, SPRAG), and (Mohler, SPRAG). Utilizing these groupings, non-contextual similarity metrics such as Jaccard, TFIDF, NegWMD, and contextual similarity metrics including USE, SBERT CE, SBERT BiE, and SimCSE in both supervised and unsupervised settings are computed for all three pairs. The ensuing outcomes are meticulously compiled within a tabulated format as specified in Table 2. Given the presence of several small values, a visual representation of these metrics is conveyed using a heatmap, depicted in the accompanying Fig. 4.

**Table. 3** Cohen's D values on the similarity scores of pairs of datasets.

| Similarity Measure | (STSB, Mohler) | (STSB, SPRAG) | (Mohler, SPRAG) |
|---|---|---|---|
| **Jaccard** | -0.482747 | -0.505967 | 0.039770 |
| **TFIDF** | -0.200502 | -0.413744 | 0.228170 |
| **NegWMD** | -0.505459 | -0.515482 | 0.020581 |
| **USE** | -0.311332 | -0.333331 | 0.012037 |
| **SBERT CE** | 0.374094 | 0.154802 | 0.283053 |
| **SBERT BiE** | 0.436234 | 0.170045 | 0.340916 |
| **SimCSE Supervised** | 0.407259 | 0.174673 | 0.295283 |
| **SimCSE Unsupervised** | 0.051829 | -0.203635 | 0.283045 |

**Fig. 5** Visual representation of Cohen's D values on pairs of datasets on the similarity metrics

For effect size quantification, the standardized mean difference using Cohen's d is computed for all the dataset pairs viz., (STSB, Mohler), (STSB, SPRAG), and (Mohler, SPRAG). The resultant values are organized within a tabular format in Table 3. As certain values exhibit positivity and a subset are relatively diminutive, visual representation in Fig. 5 aids in comprehending the effect size, facilitating a more comprehensive grasp of the outcomes.

## 5. Interpretation of Results

This section interprets the meaning of the results of experiments carried out on the three datasets. The Figures 6 and 7 illustrates a comparison between the actual similarity scores and the similarity scores calculated using non-contextual similarity metrics. In this visual representation, the color green corresponds to the Mohler dataset, yellow represents the STSB dataset, and blue pertains to the SPRAG dataset.

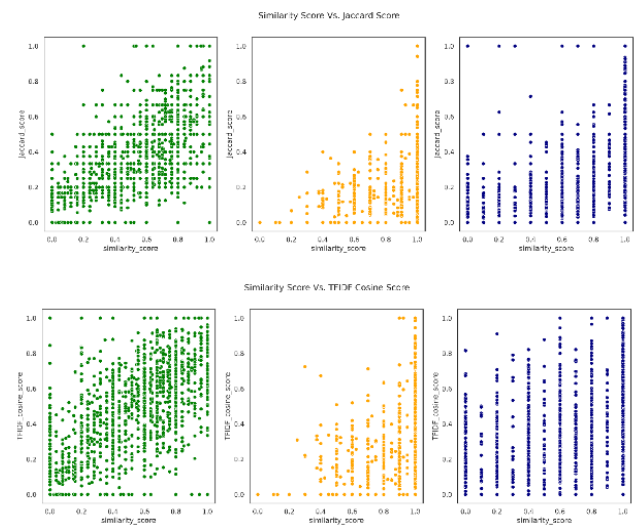

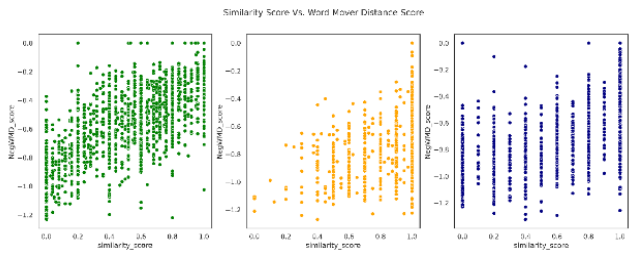

**Fig. 6.** Comparison of actual similarity_score in the dataset and the obtained non-contextual similarity scores for the three datasets viz., Mohler, STSB and SPRAG.

Among the non-contextual similarity metrics utilized, viz., Jaccard similarity, TF-IDF cosine similarity, and WMD, it is evident that these metrics yielded favorable results for the Mohler dataset, whereas their performance was less satisfactory for the STSB and SPRAG datasets. Shifting the focus to contextual similarity metrics, the figure portrays the alignment between actual similarity scores and those computed using these contextual metrics. Notably, these contextual metrics exhibited consistent performance across all datasets. They effectively captured similarity in both the Mohler and STSB datasets, yet encountered challenges with the SPRAG dataset due to its intricate sentence structures, which hindered the identification of similarity.

Analyzing the similarity scores across dataset pairs, it becomes apparent that there is a minor resemblance between the STSB and Mohler datasets, while a more pronounced similarity exists between the Mohler and SPRAG datasets. This observation is also supported by a tabulated display of the most common words across all datasets. Specifically, around 20% of the most frequently occurring words in both the Mohler and SPRAG datasets coincide. However, no common words emerge among the top 20 words in either the STSB and Mohler datasets, or the STSB and SPRAG datasets.

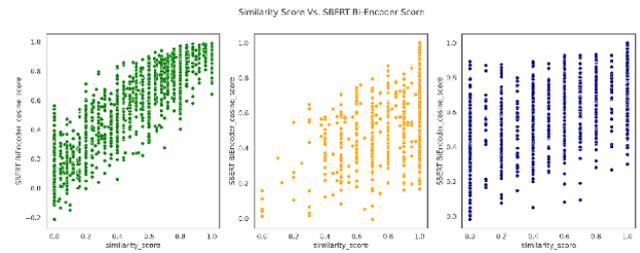

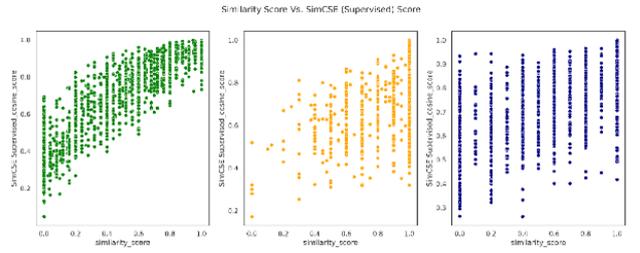

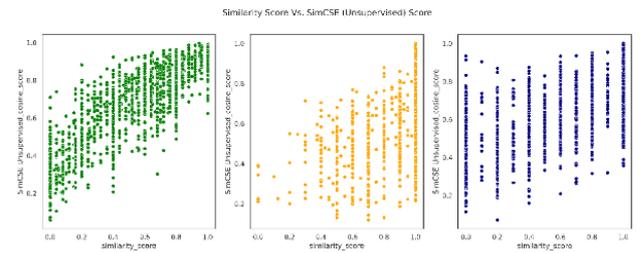

**Fig. 7.** Comparison of actual similarity_score in the dataset and the obtained contextual similarity scores for the three datasets viz., Mohler, STSB and SPRAG.

Furthermore, employing Cohen-D values to measure the effect size provides additional insight. Comparing the effect size between the SPRAG and Mohler datasets versus the SPRAG and STSB datasets, it becomes evident that the effect size is more pronounced between SPRAG and Mohler. Conclusively, leveraging the state-of-the-art models and techniques utilized for the Mohler dataset could be extended to the SPRAG dataset.

## 6. Conclusion

This work delved deeply into the domain of NLP, focusing keenly on the development of models tailored to specific datasets and their potential for transferability. The study extensively probed the feasibility of harnessing cutting-edge models, previously trained on established datasets, to achieve exceptional performance in a novel and unexplored domain. Employing a nuanced comparative analysis, which encompassed both non-contextual and contextual similarity metrics, the investigation scrutinized the intricate relationship between well-recognized benchmarks (specifically, the STSB and Mohler datasets) and a recently introduced dataset (SPRAG). Alongside, robust statistical techniques including the paired t-test and effect size were wielded to gauge the dataset relationships. The findings of the study were illuminating. They underscored the concept of transferability across NLP models. Remarkably, the analysis demonstrated that the newly introduced dataset exhibited both semantic and statistical proximity to the Mohler dataset, surpassing its similarity with the STSB dataset. This key revelation implies that the knowledge and insights accrued from the SOTA models developed for the Mohler dataset could potentially be transposed and evaluated on the novel dataset with promising outcomes.